\documentclass[12pt,a4paper]{amsart} 
\usepackage{amsfonts,amsmath,amssymb,amsthm} 
\usepackage{graphicx}
\usepackage{algorithmic}
\usepackage{tikz}
\usetikzlibrary{matrix,positioning}

\newcommand{\comment}[1]{{}}

\theoremstyle{definition}
\newtheorem{Exe}{Example} 

\begin{document} 
   
\title[V-variable image compression]{V-variable image compression
}

\author{Franklin Mendivil}
\email{franklin.mendivil@acadiau.ca}
 \address{Department of Mathematics and Statistics,
Acadia University,
Wolfville, NS B4P 2R6} 

\author{ \"{O}rjan Stenflo}
\email{stenflo@math.uu.se}

 \address{Department of Mathematics,
Uppsala University,
751 06 Uppsala,
Sweden
}


\begin{abstract}
V-variable fractals, where $V$ is a positive integer, are intuitively 
fractals with at most $V$ different ``forms'' or ``shapes''
 at all levels of magnification.
In this paper we describe how V-variable fractals can be used for the purpose of image compression.

\end{abstract} 
\maketitle

\section{introduction}

In 1988 Barnsley and Sloan \cite{BarnsleySloan88} described a method
for (lossy) digital image compression based on the idea of approximating a digital image
 with  the attractor of an iterated function system (IFS).
Their method was generalized and automated in 1992 by Jacquin \cite{Jacquin92}.  
The basic idea behind the method is to use the fact that parts of an image often resemble other parts of the image. 
Attractors of IFSs are examples of fractals. 
One drawback with using such fractals in image compression is that it restricts attention
to approximations having locally only one ``form'' or ``shape''. 
 
V-variable fractals were introduced by Barnsley, Hutchinson and Stenflo in \cite{Barnsleyetal05}, 
\cite{Barnsleyetal08}, \cite{Barnsleyetal12}.
Intuitively, a V-variable fractal is a set with at most $V$ different ``forms'' or ``shapes'' at any level of magnification.
 
The purpose of the present  paper is to describe a simple novel method for lossy  compression of digital 
images based on V-variable fractals. We make no claim that the simple 
implementation presented here
is competitive with state of the art algorithms in current use. Our purpose here is merely to announce this new approach.

The paper is organised as follows.
In Section \ref{Background} we present the mathematical background 
with definitions and basic properties of IFS attractors and V-variable fractals.
In Section \ref{method} we present our V-variable image compression method and describe how it can be automated. 
In Section \ref{compare} we compare our V-variable fractal compression 
method with the standard fractal block coding approach.
In Section \ref{general} we suggest possible generalisations of our method for future research.

\section{Background} 
\label{Background}

The present section contains  definitions and basic 
properties  of IFS attractors and V-variable fractals,
and simple examples illustrating
these objects. 

\subsection{IFS attractors}  
\label{52}

Let $(X,d)$ be a complete metric space
and let
$f_j$, $j=1,\ldots,M$ be a finite set of strict contractions on $X$, i.e.\ functions $f_j:X
\rightarrow X$, satisfying
$ d(f_j(x), f_j(y)) \leq c d(x,y) $,
 for some constant $c<1$ for any $1 \leq j \leq M$.
The set
$\{ f_j, 1 \leq j \leq M  \}$ is called an iterated function system (IFS).

From the contractivity assumption it follows that the map
\[
    \widehat{Z}({\bf i})=\lim_{n \rightarrow \infty}  f_{i_1} \circ f_{i_2} \circ \cdots \circ f_{i_n}(x_0),
\] 
exists for any  ${\bf i}=i_1 i_2 \ldots \in \{1,\ldots,M\}^{\mathbb
  N}$ and the limit is independent of $x_0 \in X$.
The set of all limit points 
\[
   A=\{\widehat{Z}({\bf i}): \ {\bf i} \in \{1,\ldots,M\}^{\mathbb N} \} \subseteq X
\]
is called the attractor of the IFS.

If $A_0 \subseteq X$ is a compact subset of $X$ such that $f_i (A_0) \subseteq A_0$, for all $1\leq i \leq M$, and 
 $A_n= \cup_{i_1,\ldots,i_n} f_{i_1} \circ f_{i_2} \circ \cdots \circ f_{i_n}(A_0)$,
where the union is taken over all indices $(i_1,\ldots,i_n) \in \{1,\ldots,M\}^n$,
then $A_{n+1}=\cup_{i=1}^M f_i(A_n) \subseteq A_n$, for all $n$, and $A = \cap_{n=1}^\infty A_n$.

\begin{Exe}
\end{Exe}
The (middle-third) Cantor set is the attractor of the IFS with functions
 $f_1(x)=x/3$ and $f_2(x)=x/3+2/3$.
The Cantor set is an example of a self-similar fractal.  

Let $A_0=[0,1]$. If $A_n:= \cup_{i_1,\ldots,i_n} f_{i_1} 
\circ f_{i_2} \circ \cdots \circ f_{i_n}(A_0)$  then $A_n$ is a union of 
$2^n$ intervals all of length $(1/3)^n$, and the (middle-third) Cantor
 set $A$ is the limiting set of the sequence of sets $\{A_n\}$ as $n \rightarrow \infty$. 
Any fixed interval  $f_{i_1} \circ f_{i_2} \circ \cdots \circ f_{i_n}(A_0)$ of
 $A_n$ will contain the two intervals
 $f_{i_1} \circ f_{i_2} \circ \cdots \circ f_{i_n}(f_1[0,1])$ and
 $f_{i_1} \circ f_{i_2} \circ \cdots \circ f_{i_n}(f_2[0,1])$ of
 $A_{n+1}$. Visually this property corresponds to ``deleting the middle third piece'' of each interval in 
$A_n$ in order to obtain $A_{n+1}$ from $A_n$.

\begin{center} 
   \resizebox{!}{40mm}{\includegraphics{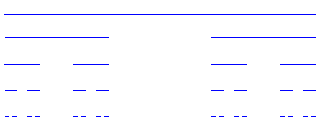}} \label{Cantor}
\end{center}
\hspace{23mm}\parbox{12cm}{\em  The first $5$ approximating sets $A_0$, $A_1$
$A_2$, $A_3$, $A_4$ of  the limiting Cantor set $A$.  }

$$$$

A natural generalisation of the above Cantor-set construction  is to delete different proportions of all 
intervals involved in the construction. 
This corresponds to using  different  IFSs  controlling how much we delete for different intervals of $A_n$.
Such a generalisation leads to the concept of code tree fractals:

\subsection{Code tree fractals}

Let $ \{f_j^\lambda, 1 \leq j \leq M    \}_{\lambda \in \Lambda}$, be an indexed family of IFSs, where
$f_j^\lambda: X \rightarrow X$, are strict contractions on a complete metric space $(X,d)$, 
$M$ is a finite positive integer and  $\Lambda$ is a  finite index set.

Consider a function $\omega\colon  \bigcup_{k=0}^\infty \{ 1,\ldots,M \}^k \rightarrow \Lambda$.
We call $\omega$  a code tree. 
A code tree can be identified with a labelled infinite  $M$-ary tree with each node labelled with the index of an IFS, some $\lambda \in \Lambda$.

Define 
\[
    \widehat{Z}^{\omega}({\bf i})= 
        \lim_{k \rightarrow \infty} f_{i_1}^{\omega(\emptyset)}
     \circ
      f_{i_2}^{\omega(i_1)} \circ\dots\circ  f_{i_k}^{\omega(i_1\ldots i_{k-1})}(x_0), \ 
          \text{for} \ {\bf i}\in  \{ 1,\ldots ,M \}^{\mathbb{N}},  
\]
and 
\[ 
    A^\omega = \{ \widehat{Z}^\omega({\bf i}) ; \ {\bf i}\in  \{ 1,\ldots,M \}^{\mathbb{N}} \},
\]
for some fixed $x_0 \in X$. (It doesn't matter which $x_0$ we choose, since the limit is, as before, independent of $x_0$.)
We call $A^\omega$  the attractor or code tree fractal corresponding to the code tree $\omega$ and will refer to  ${\bf i}$ as an address of the point $\widehat{Z}^\omega({\bf i})$ on  $A^\omega$.

Let $A_0 \subseteq X$ be a compact subset of $X$ such that $f_j^\lambda (A_0) \subseteq A_0$, for all 
$1\leq i \leq M$, and $\lambda \in \Lambda$.
Let 
   $A_n^\omega= \cup_{i_1,\ldots,i_n} f_{i_1}^{\omega(\emptyset)} \circ
           f_{i_2}^{\omega(i_1)} \circ\dots\circ  f_{i_n}^{\omega(i_1\ldots i_{n-1})}(A_0)$,
where the union is taken over all indices $(i_1,\ldots,i_n) \in \{1,\ldots,M\}^n$. 
Any fixed subset 
 $f_{i_1}^{\omega(\emptyset)}  \circ  f_{i_2}^{\omega(i_1)} \circ\dots\circ  f_{i_n}^{\omega(i_1\ldots i_{n-1})}(A_0)$
of
 $A_n^\omega$ contains the $M$ sets 
 \[
     f_{i_1}^{\omega(\emptyset)} \circ
                 f_{i_2}^{\omega(i_1)} \circ\dots\circ  f_{i_n}^{\omega(i_1\ldots i_{n-1})}(f_j^{\omega(i_1\ldots i_{n})}(A_0)), \quad j=1,\ldots ,M
\]
 of 
$A_{n+1}^\omega$.  
It follows that
 $A_{n+1}^\omega \subseteq A_n^\omega$, for all $n$, and $A^\omega = \cap_{n=1}^\infty A_n^\omega$.

\begin{Exe} \label{codecantor}  
\leavevmode
\smallskip


\begin{center} 
  \includegraphics[width=4 in, height = 1 in]{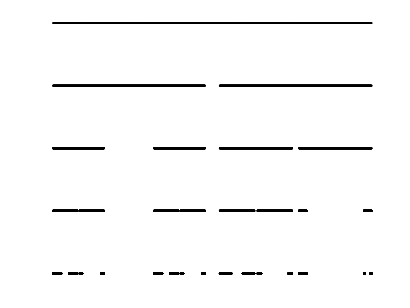}
  \label{twovarcantor}
\end{center}

\medskip

The first $5$ approximating sets $A_0$,
 $A_1^\omega$
$A_2^\omega$, $A_3^\omega$, $A_4^\omega$ of  a  limiting code tree fractal 
$A^\omega$ generated by the IFSs 
$\{f_1^1(x)=10x/21,f_2^1=10x/21+11/21\}$,
$\{f_1^2(x)=x/3,f_2^2=x/3+2/3\}$, $\{f_1^3(x)=x/10,f_2^3=x/10+9/10\}$ 
and the code tree with first 3 levels given by

\begin{center}
   \includegraphics[width = 3 in]{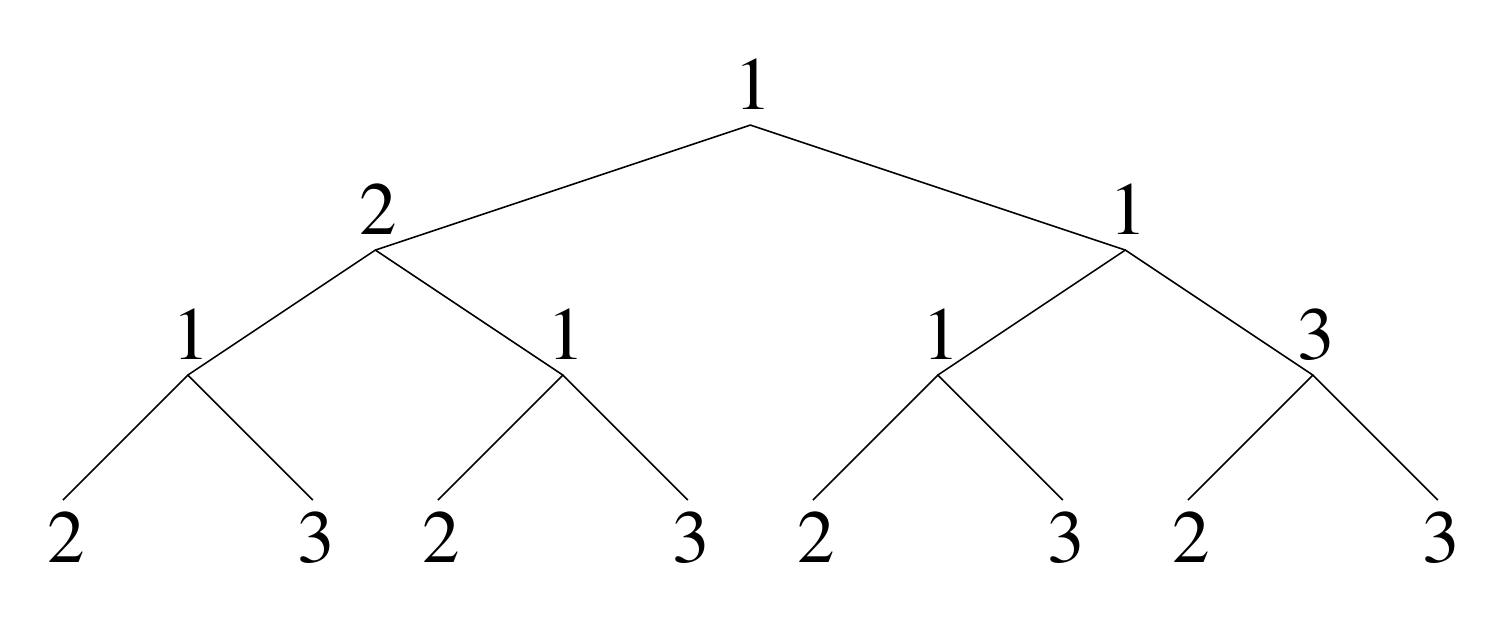}
\end{center}

Intuitively the 3 IFSs corresponds to ``cut a small piece'', ``cut a middle third piece''
 and ``cut a big piece'' respectively, in each step  of the construction.
\end{Exe}

\noindent
{\bf V-variable fractals:} \\
The sub code trees of a code tree $\omega$ corresponding to a node 
$i_1\ldots i_k$ is the code tree $\omega_{i_1\ldots i_k}$ defined by
 $\omega_{i_1\ldots i_k}(j_1j_2\ldots j_n):=\omega(i_1\ldots i_k j_1\ldots j_n)$, for any $n\geq 0$ and $j_1\ldots .j_n \in  \{1,\ldots ,M\}^n$.

Let $V \geq 1$ be a positive integer. We call a code tree V-variable if for any $k$ the set of code trees $\{  \omega_{i_1\ldots i_k}; i_1\ldots i_k \in  \{1,\ldots ,M\}^k \}$ contains at most $V$ distinct elements.

A code tree fractal $A^\omega$ is said to be V-variable if $\omega$ is a V-variable code tree.

A V-variable fractal is intuitively a fractal having at most $V$ distinct ``forms'' or ``shapes'' at any level of magnification.\\

\noindent
{\bf Example 2 (continued):} \label{121019}
The code tree in Example  \ref{codecantor} is
 2-variable (and thus V-variable for any $V\geq 2$) up to level 3.
At level 0 there is only one sub code tree (the code tree itself).
At level 1 the sub code trees (up to level 2) are given by
\begin{center}
     \includegraphics[width = 1.5 in]{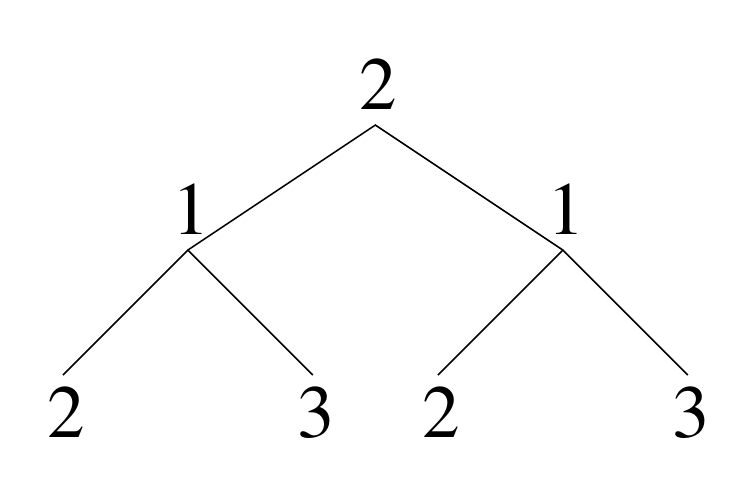} \hskip 0.4 cm  \raisebox{1 cm}[\height][\depth]{\mbox{ and }}  \hskip 0.4 cm  \includegraphics[width = 1.5 in]{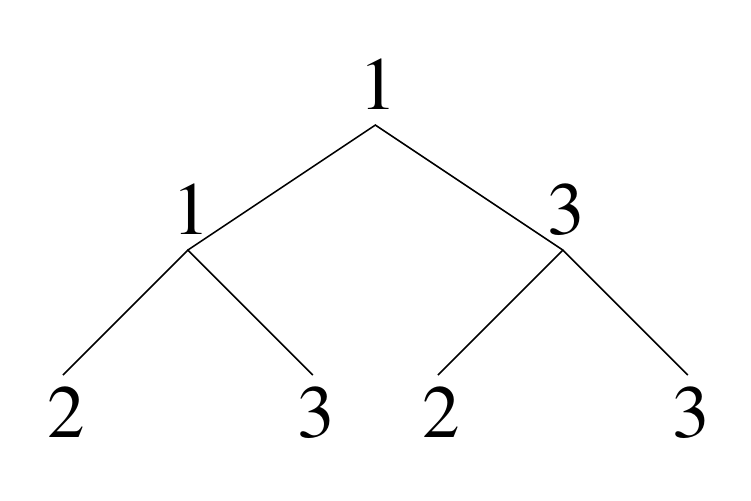}
\end{center}
At level 2 the distinct sub code trees (up to level 1)
 are given by
\begin{center}
     \includegraphics[width = 0.8 in]{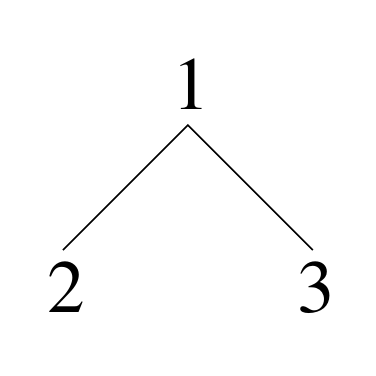} \hskip 0.4 cm  \raisebox{1 cm}[\height][\depth]{\mbox{ and }}  \hskip 0.4 cm  
\includegraphics[width = 0.8 in]{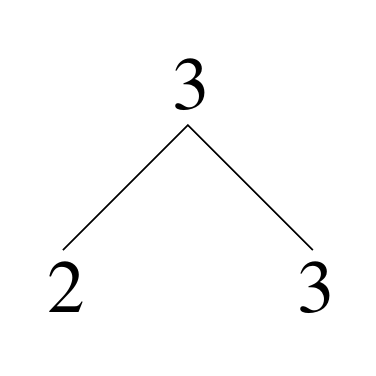}
\end{center}
and at level 3 the 2 distinct sub code trees (up to level 0) are given by 2 and 3.

The V-variable structure of a V-variable code tree can be described by a
V-variable ``skeleton tree'' where nodes are labelled according to which of the $V$  types
the sub code tree rooted in the given node belongs, where the sub code trees are labelled in order 
of appearance from left to right.\\

\noindent
{\bf{Example \ref{codecantor}  (continued):}}
The code tree in  Example \ref{twovarcantor} has a skeleton tree (up to level 3) given by 

\begin{center}
   \includegraphics[width = 3 in]{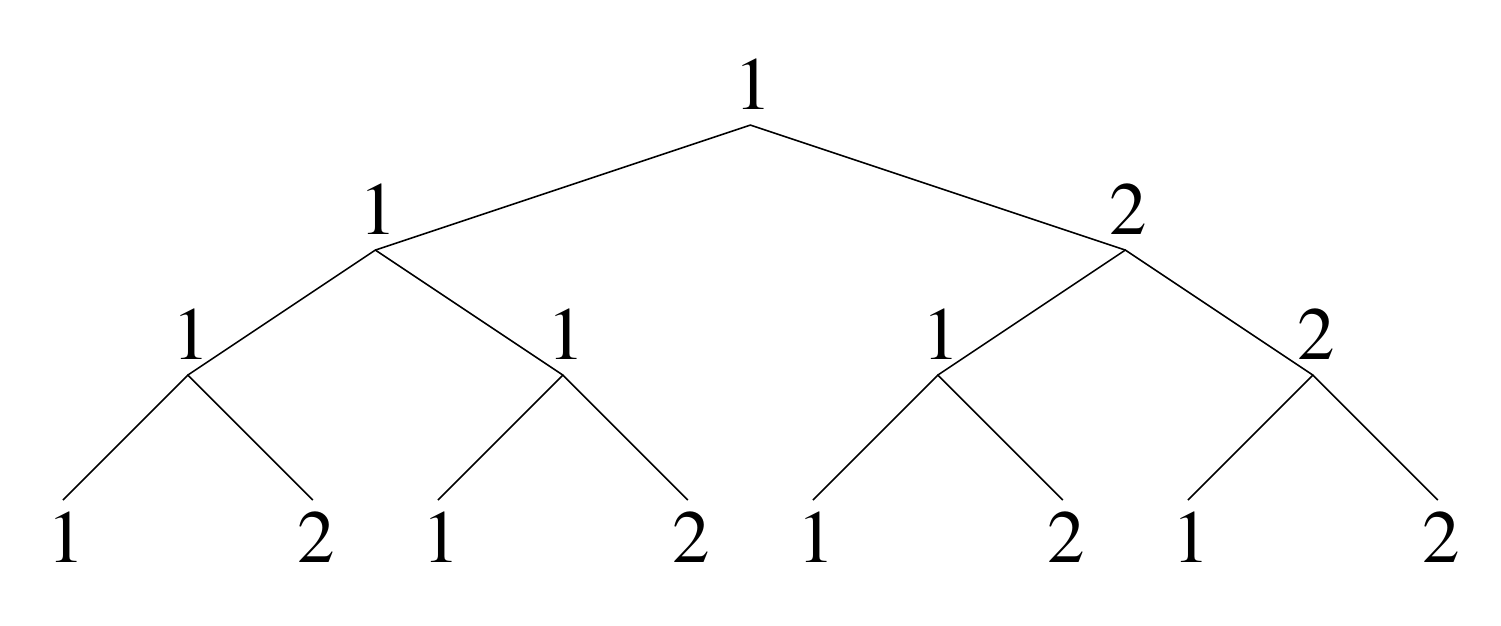}
\end{center}

We can describe the skeleton tree up to level $n$ by a $(V \cdot M) \times n$ matrix, where the $k$th column specifies the labels of the ordered $M$ ``child'' nodes at level $k$ for each of the ordered $V$ possible ``parental'' node types at level $k-1$, $k = 1, 2, \ldots, n$.
Recall that $M$ is the number of functions per IFS.

Thus given a skeleton tree of a V-variable code tree (up to level $n$), we can represent the skeleton tree by a $(V \cdot M) \times n$ matrix
where the element on row $(L-1) \cdot M+i_k$ and column $k$ gives the label of $i_1,\ldots ,i_{k}$ if the label of 
$i_1,\ldots ,i_{k-1}$ is $L \in \{1,\ldots ,V\}$. 

\noindent
{\bf{Example \ref{codecantor}  (continued):}}
The skeleton-tree in Example \ref{codecantor} above can be represented by
\[
  \begin{bmatrix} 
 1 & 1 & 1 \\
 2 & 1 & 2 \\ 
 \cdot & 1 & 1 \\
 \cdot & 2 &  2
\end{bmatrix}.
\]
(We can thus store the first $n$ levels of a V-variable skeleton tree with much less information than
the storage of the first $n$ levels of a general code tree if $n$ is large.
This property will be used in a crucial manner in our V-variable image compression algorithm to be described below.)\\

In order to generate a V-variable code tree from the V-variable skeleton tree (up to level $n$),
we need  labelling functions $Q_k : \{ 1,\ldots ,V\} \rightarrow \Lambda$, for each level $0 \leq k \leq n$. 
A node on level $k$ with label $j$ in the skeleton-tree is labelled by $Q_k(j)$ in the code tree.
Given a maximum  level $n$, we may represent the functions $Q_0,\ldots ,Q_n$ with an 
$(n+1) \times M$ matrix $Q$ where $Q(i,j)= Q_{i-1}(j)$, so in Example
\ref{codecantor} we would get
\[
 Q=
 \begin{bmatrix}
  1 &  \cdot \\ 
  2 & 1 \\ 
  1 & 3 \\ 
  2 & 3 
 \end{bmatrix}.
\]

All IFS attractors can be regarded as being $1$-variable fractals. See e.g.\ Barnsley
et al.\  \cite{Barnsleyetal05}, \cite{Barnsleyetal08} and  \cite{Barnsleyetal12} 
for more on the theory of V-variable fractals. \\

\noindent
{\bf{Coloured V-variable fractals and images:}}
A simple way to colour a V-variable fractal is to  assign colours using its V-variable structure.

\begin{Exe} \label{ex:3}
 Any V-variable fractal generated by the single IFS \\
\[
\begin{array}{ll}
  \Big\{ 
  f_1^1  \! \!  \begin{pmatrix} x \\  y  \end{pmatrix} 
     \hspace{-1mm} = \hspace{-1mm}
   \begin{pmatrix} 1/2 & 0 \\ 0 & 1/2  \end{pmatrix} 
     \begin{pmatrix} x \\ y  \end{pmatrix}, & \hskip 1 cm
  f_2^1 \begin{pmatrix} x \\  y  \end{pmatrix} 
   \hspace{-1mm} = \hspace{-1mm}
    \begin{pmatrix} 1/2 & 0 \\ 0 & 1/2  \end{pmatrix}
    \begin{pmatrix} x \\  y  \end{pmatrix} 
   \hspace{-1mm} + \hspace{-1mm}
     \begin{pmatrix} 0 \\ 1/2 \end{pmatrix},\cr \cr
 f_3^1 \begin{pmatrix} x \\  y  \end{pmatrix} 
   \hspace{-1mm} = \hspace{-1mm}
     \begin{pmatrix} 1/2 & 0 \\  0 & 1/2  \end{pmatrix}
      \begin{pmatrix} x \\  y  \end{pmatrix} 
 \hspace{-1mm} + \hspace{-1mm} 
    \begin{pmatrix} 1/2 \\  0  \end{pmatrix}, & \hskip 1 cm
 f_4^1 \begin{pmatrix} x \\  y  \end{pmatrix} 
   \hspace{-1mm} = \hspace{-1mm}
 \begin{pmatrix} 1/2 & 0 \\  0 & 1/2 \end{pmatrix}
 \begin{pmatrix} x \\  y \end{pmatrix} 
 \hspace{-1mm} + \hspace{-1mm}
   \begin{pmatrix} 1/2 \\ 1/2  \end{pmatrix} \Big\},   \cr
 \end{array}                      
\]
will be the unit square (the attractor of the IFS), so a given V-variable skeleton tree plays no role when characterizing this set.
(Note that $Q(i,j)=1$ for all $i,j$ here).
 
The unit square can be regarded as being built up from $4^n$ disjoint smaller squares of size 
$(1/2)^n \times (1/2)^n$, where any given square is associated to one out of $V$ distinct types using the given V-variable skeleton-tree.
A simple way to colour the unit square is therefore to first choose an arbitrary $n$
and then colour each of its $(1/2)^n \times (1/2)^n$ squares depending on its type. 
By  identifying the unit square with a 
rectangular  $512 \times 512$ pixel-``computer screen,'' using $n=9$ corresponds to colouring pixels
using $V$ different colour values, see Example \ref{4varEx} below.   
\end{Exe}

A digital $j \times k$  (8-bit) grayscale image consists of $j \cdot k$ pixels where each pixel is assigned a 
pixelvalue in $ \{0,1,2,\ldots ,255\}$.  
In order to avoid blurring the exposition we will, for convenience, only consider $512 \times 512$ images in this paper.  
For any $0 \leq n \leq 9$, we may divide a given $512 \times 512=2^9 \times 2^9$ image  into
$4^n$ nonoverlapping  $2^{9-n} \times 2^{9-n}$ pieces.
We call these pieces the image pieces of generation or level $n$.

\begin{figure}[hp]
\begin{center}
\begin{tabular}{cc}
\includegraphics[width=2.0in]{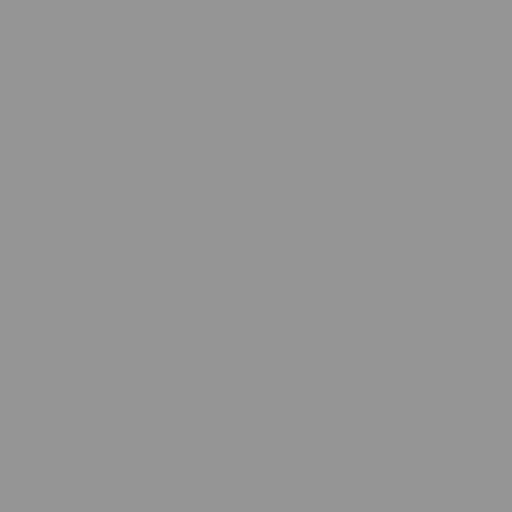} &
\includegraphics[width=2.0in]{Enya4} \\ 
\includegraphics[width=2.0in]{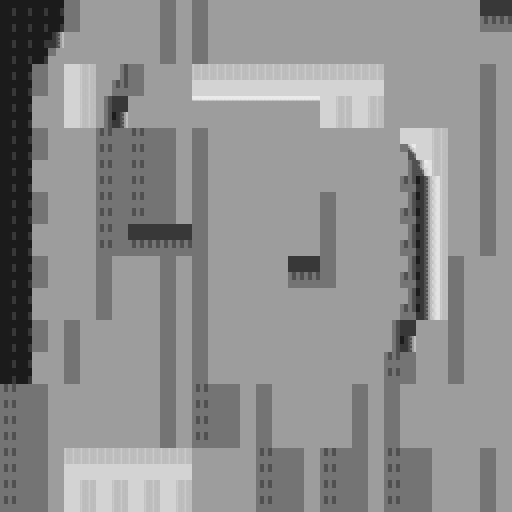} &
\includegraphics[width=2.0in]{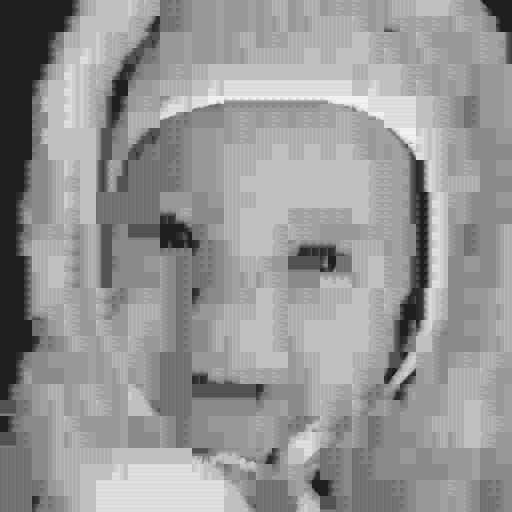} \\
\includegraphics[width=2.0in]{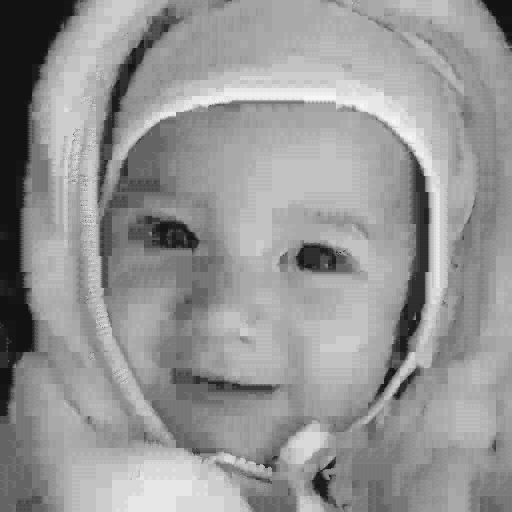} &
\includegraphics[width=2.0in]{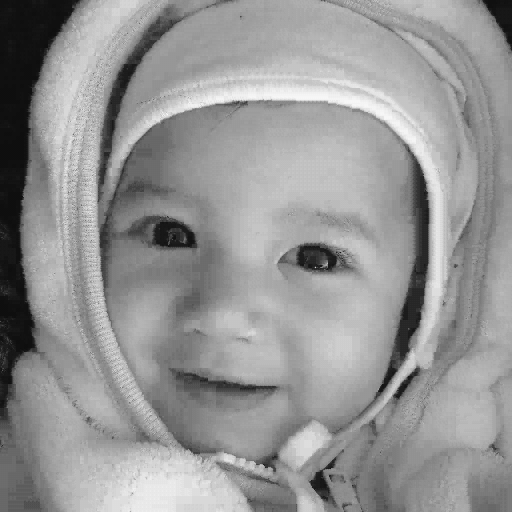} \\
\end{tabular}
\caption{1-variable, 4-variable, 16-variable,  64-variable, 256-variable and 1024-variable approximations of our $512 \times 512$ grayscale test image. 
We can store an  $4^n$ variable approximation, with  $1 \le n \le 8$, using  $(8-n)4^{n+1}$ numbers in $\{1,\ldots ,4^n\}$ 
(each such number stored in $2n$ bits) plus $4^{n+1}$ numbers in $\{0,1,\ldots ,255\}$  (each such number stored in 1 byte=8 bits).  
Thus we can store the $4^n$-variable approximation in $4^{n+1}(2n(8-n)+8)/8=4^n( n(8-n)+4)=4^n(20-(n-4)^2)$ bytes, 
so the images above are stored in $1$B, $44$B, $256$B, $1216$B, $5120$B, and $19456$B for $n=0,1,2,3,4,5$ respectively.  
The original picture is stored with $512 \cdot 512=262144$B (since each pixel is assigned a number in  $\{0,1, \ldots, 255\}$ 
and thus requires one byte of storage. In particular it therefore follows that the original image are stored with a compression 
ratio of $512 \cdot 512/1216\approx 215.6$,  $(512 \cdot 512)/5120=51.2$, and $(512 \cdot 512)/19456 \approx 13.5$ for the
 last 3 images above. Note that the storage space can be  reduced with increased compression ratios if we apply 
some lossless  compression technique (such as for example entropy coding).}
 \label{fig:approxfigurs}
\end{center}
\end{figure}

\begin{Exe} \label{4varEx} 
The 4-variable $512 \times 512$ grayscale image


\begin{center}
\includegraphics[width=1.5in]{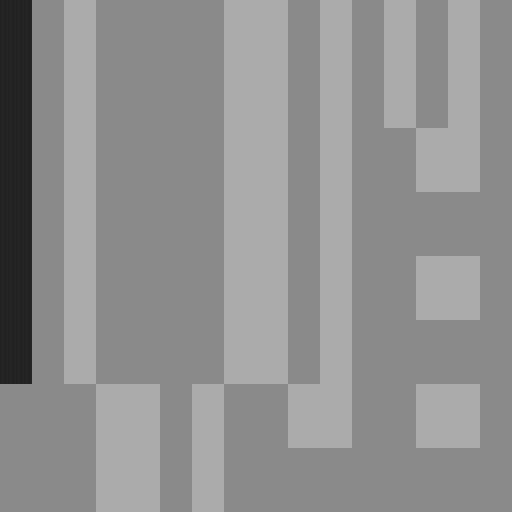} \\ 
\end{center}


\noindent can be built up by $4^n$ image pieces of size 
$2^{9-n} \times 2^{9-n}$ of (at most) $4$ distinct types, for any
 $n=0,1,2,\ldots ,9$.
The appearance of these image pieces depends on $n$.
If e.g.\  $n=2$ then the $16$  image pieces of size $128 \times 128$ pixels are of the 4 types
\begin{center} 
    \resizebox{!}{14mm}{\includegraphics{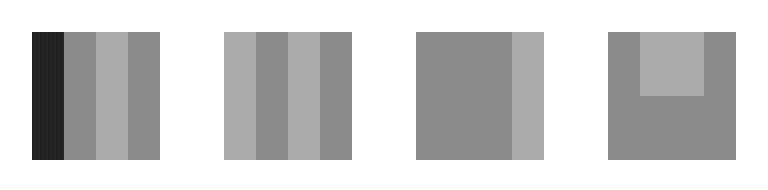}} \label{Enya4_4_montage}
\end{center}
and if $n=3$ then  the $64$  image pieces of size $64 \times 64$ pixels
are of the 4 types
\begin{center} 
   \resizebox{!}{7mm}{\includegraphics{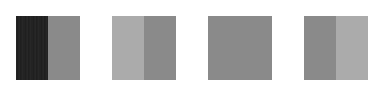}} \label{Enya4_8_montage}
\end{center}

\noindent
By looking at the image, and its image pieces,  we see that we can, recursively, describe 
the $4$-variable image using $4$ images of smaller and smaller size, i.e.\ recursively describe more and more levels of the V-variable skeleton tree of the image:  

\begin{center} 
   \resizebox{!}{100mm}{\includegraphics{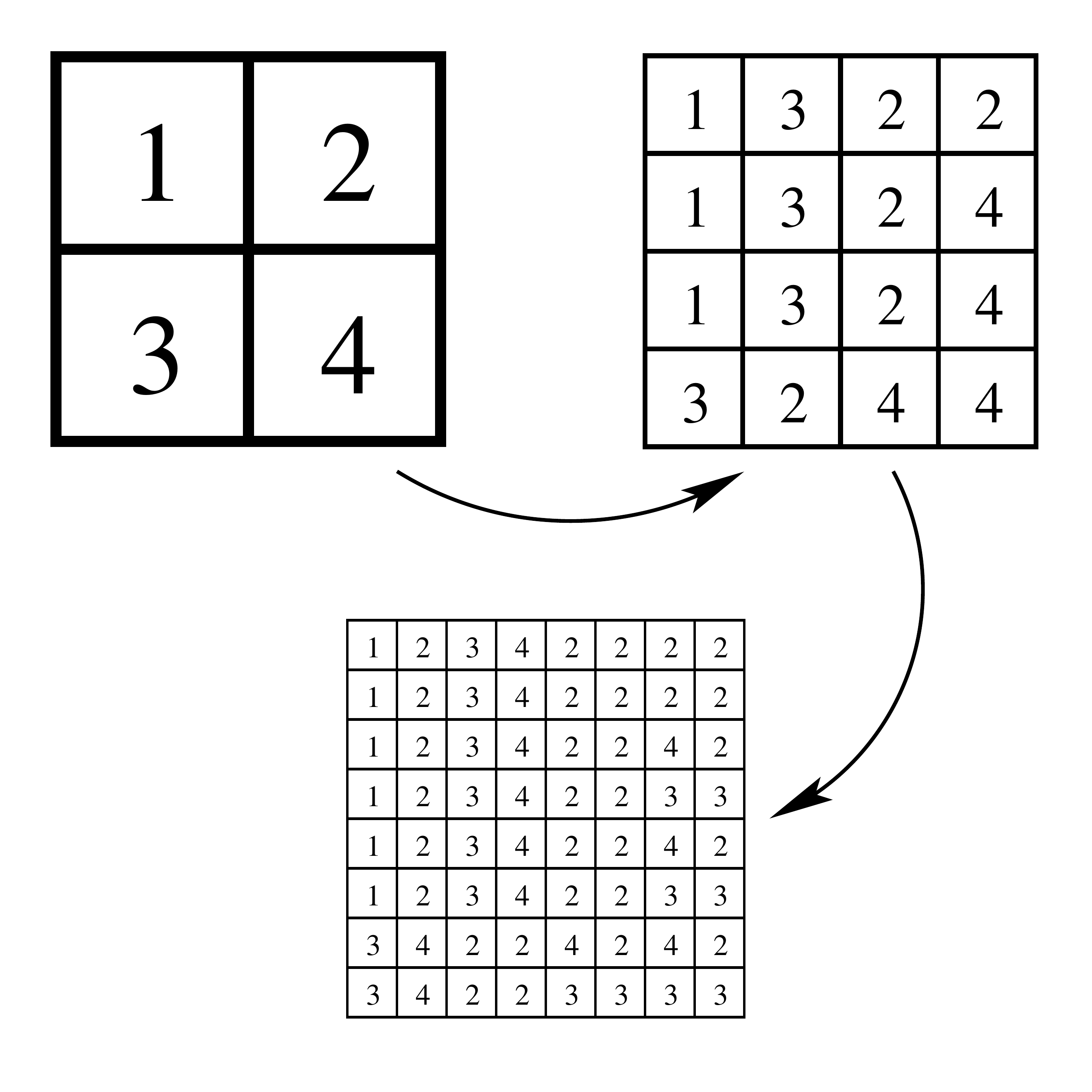}} 
\end{center}

At each stage, we replace a block of the current level with four blocks from the next level.   There are (at most) $V$ types of blocks at each level and the substitution is done according to the type.  For example, we see that in the second stage (illustrated in the figure above),  all blocks of type 3 are replaced by the same thing (a block with numbers 3,4,3,4).  The first two (non-trivial) steps as shown above can be visually described by the substitutions:

\begin{center} 
   \resizebox{!}{60mm}{\includegraphics{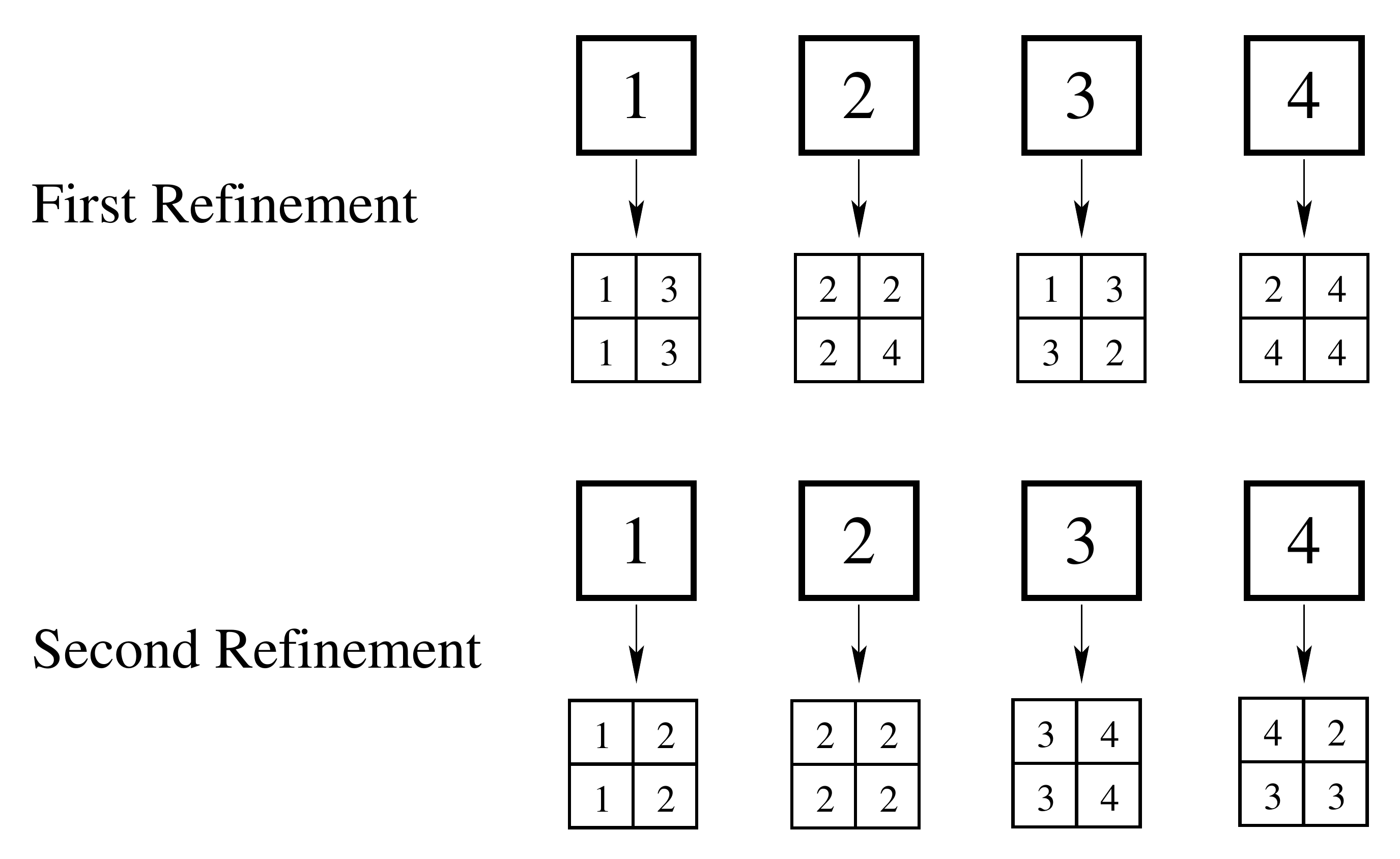}} 
\end{center}

\noindent
A full characterization of  the image  (See also the second image in Figure \ref{fig:approxfigurs})
is given by the $16 \times 8$ matrix
\[ 
\begin{pmatrix}
          1 &    1  &   4 &    3  &   3  &   3   &  2 &  138 \\
          3  &   2 &    1  &   2  &   3  &   3   &  4  & 138 \\
          1  &   1  &   4 &    3  &   3  &   3 &    2  & 138 \\
          3   &  2  &   1 &    2  &   3  &   3   &  4 &  138 \\
           2  &   2 &    2 &    4  &   1 &    4   &  3 &   33 \\
          2  &   2  &   3  &   4  &   1 &    4  &   3 &   33 \\
          2  &   2 &    2  &   4    & 1&     4    & 3  &  33 \\
          4  &   2  &   3 &    4    & 1 &    4 &    3   & 33 \\
          1   &  3 &    1  &   2   &  1  &   1  &   1  & 171 \\
          3   &  4 &    1  &   2  &   1   &  1  &   1 &  171 \\
          3   &  3 &    1  &   2 &    1  &   1   &  1  & 171 \\
          2   &  4 &    1 &    2  &   1  &   1 &    1 &  171 \\
          2   &  4 &    3    & 1 &    2 &    2    & 3 &   37 \\
          4  &   2 &    2 &     1 &    4 &    2  &   3 &   37 \\
          4   &  3 &    3 &    1 &    2   &  2   &  3  &  37 \\
         4  &   3 &    2 &    1 &    4 &    2 &    3  &  37 \\
\end{pmatrix}.
\]
Each column stores one (non-trivial) level in the V-variable skeleton tree of the image.
(The previous image corresponds to the first two columns.)
In the last column we store the $V=4$ different grayscale values used for all pixels.
Visually it is hard to see more than 3 colours in the image since the grayscale values of 33 and 37 look almost the same.
In general we can store a $4^j$-variable grayscale image using a $(4 \cdot 4^j) \times (9-j)$ matrix. 

\end{Exe}

\section{V-variable image compression} 
\label{method}

We will restrict attention to two-dimensional grayscale images here  
for convenience. Generalisations of our method to higher dimensions and  
color images are straightforward.

The idea of our method is to approximate a given image with a V-variable
image with the property that we have at most $V$ distinct image pieces in generation $n$, 
for any $0\leq n \leq 9$, where $V$ is a given positive integer.

\subsection{Description of the V-variable image compression algorithm} 

Let $1 \le V < 4^8$ be a fixed integer. We can find a V-variable approximation of the given image by using 
the following algorithm:

\bigskip

\begin{algorithmic}[1]

     \STATE   Find the level $n_0 \ge 0$ so that $4^{n_0} \le V < 4^{n_0 + 1}$.  Take all the $4^{n_0}$ image pieces on the $n_0$th level to be distinct.
 
      \STATE   Classify each of the $4^{n_0+1}$ image pieces of level $n_0 + 1$ into $V$ clusters and identify each cluster with a representative image.

     \FOR{Level $n$ from $n_0 + 2$ to $9$}

          \STATE  Given the $V$ cluster representatives of level $n-1$, divide each of these $V$ images into $4$ distinct images for level $n$.
Classify these $4V$ images into $V$ clusters and identify each cluster with a representative image for level $n$.

     \ENDFOR

      \STATE  For level $n = 9$, the representative images will be ``one pixel images'' which can be identified with a value in $\{0,1,\ldots, 255\}$.  Any ``one pixel image'' is identified with the corresponding cluster value.

\end{algorithmic}

\subsection{Storage requirements}

The classification of clusters requires $4^{n_0 + 1}$ numbers in $\{1,\ldots, V\}$ for level $n_0 + 1$, $0 \leq n_0 \le 7$, and $4 V$ numbers in $\{1,\ldots, V\}$ for levels
$n_0 + 2 \le n \le 8$.  For level $9$ we identify the cluster and image representatives with pixelvalues and we therefore need $4 V$ numbers in $\{ 0, \ldots, 255 \}$
for level $n = 9$ and no further information to store the image representatives.

Thus, for $V>1$, $0 \le n_0 \le 7$, our V-variable method of storing requires in total  $4^{n_0+1}+4V (7-n_0)$ numbers in $\{1,\ldots ,V\}$ plus
 $4 V$ numbers in $\{0,\ldots ,255\}$, where $4^{n_0} \leq V < 4^{n_0+1}$. 

In the special case when $V = 4^{n_0}$ then it is convenient to store the code in an $4V \times (9 - n_0)$ matrix \emph{Code} with values in $\{1,\ldots, V\}$, except for the last
column with values in $\{0,\ldots, 255\}$, where the columns successively correspond to information of the V-variable skeleton-tree for levels $n_0+1,\ldots, 9$, 
see the end of Example \ref{4varEx}   and Section
\ref{subsec:reconstruction}.

\subsection{Automating the V-variable image compression algorithm in Matlab}

The main tool needed in order to automate the algorithm above is a way to classify $n$ images into $k$ clusters. 
Such a clustering can be done in many different ways. For simplicity, in our implementation below we have chosen to use Matlab's built-in command
{\tt kmeans}.  The K-means algorithm is a popular and basic clustering algorithm which  finds clusters and cluster representatives for a set of vectors by iteratively minimizing
the sum of the squares of the ``within cluster''  distances (the distances from each of the vectors to the closest cluster representative) \cite{HTF}.  We  treat the sub-images as vectors and use the standard Euclidean distance to measure similarity.  We also use random initialization of the cluster representatives.

\subsection{Reconstruction of an image from its code}
\label{subsec:reconstruction}

The process of reconstructing an image based on its V-variable code is quick.
The address of a pixel in an $512 \times 512$ image can be described
by a sequence $i_1 i_2\ldots i_9 \in \{1, 2,3,4\}^9$ by identifying the $512 \times 512$ grid with the unit square, as in Example \ref{ex:3}.
Below we illustrate how to calculate the value of a pixel with address
$322113414$ for the $4$-variable approximation of our test image:

\bigskip

\begin{center}

\begin{tikzpicture}[node distance= 10 ex]
  \matrix (mymatrix) [matrix of math nodes,left delimiter={(},right delimiter={)},column sep = 15 pt]
  {
          1 &    1  &   4 &    3  &   3  &   3   &  2 &  138 \\
          3  &   2 &    1  &   2  &   3  &   3   &  4  & 138 \\
          1  &   1  &   4 &    3  &   3  &   3 &    2  & 138 \\
          3   &  2  &   1 &    2  &   3  &   3   &  4 &  138 \\
           2  &   2 &    2 &    4  &   1 &    4   &  3 &   33 \\
          2  &   2  &   3  &   4  &   1 &    4  &   3 &   33 \\
          2  &   2 &    2  &   4    & 1&     4    & 3  &  33 \\
          4  &   2  &   3 &    4    & 1 &    4 &    3   & 33 \\
          1   &  3 &    1  &   2   &  1  &   1  &   1  & 171 \\
          3   &  4 &    1  &   2  &   1   &  1  &   1 &  171 \\
          3   &  3 &    1  &   2 &    1  &   1   &  1  & 171 \\
          2   &  4 &    1 &    2  &   1  &   1 &    1 &  171 \\
          2   &  4 &    3    & 1 &    2 &    2    & 3 &   37 \\
          4  &   2 &    2 &     1 &    4 &    2  &   3 &   37 \\
          4   &  3 &    3 &    1 &    2   &  2   &  3  &  37 \\
         4  &   3 &    2 &    1 &    4 &    2 &    3  &  37 \\
  };
\node [left] at (-6,4.1) {$1$};
\draw [blue] (-6.25,4.1) circle (8 pt);
\draw (-6,3.9) -- (-5.5,3.2);
\node [below,blue] at (-5.9,3.65) {$3$};

\node [left] at (-5,4.1) {$1$};
\node [left] at (-5,3.55) {$2$};
\node [left] at (-5,3.0)  {$3$};
\node [left] at (-5,2.45) {$4$};

\draw [blue] (-5.25,3) circle  (8 pt);
\draw (-5.1,2.7) -- (-4.0,-0.70);
\node [left,blue] at (-4.6, 1.0) {$2$};

\draw [blue] (-3.75, -0.8) circle (8 pt);

\draw (-3.4,-0.80) -- (-3.1,-0.80);
\draw [blue] (-2.75,-0.8) circle (8 pt);

\node [below,blue] at (-3.2, -0.85) {$2$};

\draw (-2.50, -1.1) -- (-1.9, -2.1);
\draw [blue] (-1.7,-2.4) circle (8 pt);
\node [below,blue] at (-2.25,-1.65) {$1$};

\draw (-1.5,-2.1) -- (-0.95,-0.55);
\draw [blue] (-0.72,-0.25) circle (8 pt);
\node [above,blue] at (-1.35,-1.4) {$1$};

\draw (-0.45,-0.0) -- (0.0,0.6);
\draw [blue] (0.29,0.82) circle (8 pt);
\node [above,blue] at (-0.29, 0.25) {$3$};

\draw (0.5,1.1) -- (1.1,2.2);
\draw [blue] (1.3,2.5) circle (8 pt);
\node [above,blue] at (0.63,1.5) {$4$};

\draw (1.55,2.2) -- (2.1,0);
\draw [blue] (2.3,-0.25) circle (8 pt);
\node [below,blue] at (1.65,0.8) {$1$};

\draw (2.5,-0) -- (3.2,2.2);
\draw [blue] (3.55,2.5) circle (11 pt);
\node [below,blue] at (2.9,0.8) {$4$};

\end{tikzpicture}

\end{center}

\bigskip

\noindent
By default the label of the whole unit square is $1$. 
Now proceed inductively; If the label of the square with address starting with
$i_1 \ldots i_{k-1}$  is $L \in \{1,\ldots ,V\}$, for $1 \leq k \leq 9$,  
then we label the square with address starting with  $i_1\ldots i_{k}$   by the element on row $4(L-1) +i_k$ and column $k$ in the $(4 \cdot V) \times 10$ coding matrix
extended with the trivial coding, i.e. in the matrix $Q$ 
defined by
$Q(i,j)=i$, if $i \leq 4^j$, and $0 \leq j \leq n_0$, and
$Q(i,j) = \mbox{\emph{Code}}(i,j-n_0)$, if $n_0 < j \le 9$.

\section{Description and comparison with the standard fractal block method} 
\label{compare} 


In this section we describe the fractal block coding algorithm (developed by Jacquin in \cite{Jacquin92}) and give a brief comparison with our method.
There are many different variations; we describe the most basic here.

\begin{figure}[h]
  \begin{center}
    \includegraphics[width=2in, height=2in]{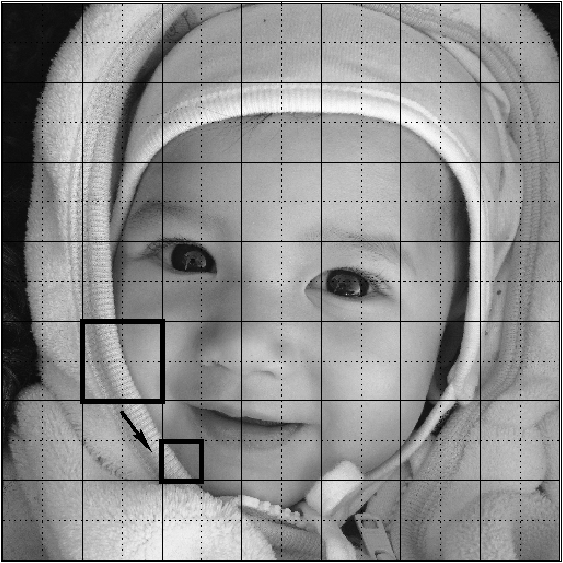}
    \caption{Basic block decomposition and mapping a large block to a smaller block.}
    \label{fig:block_ifs}
  \end{center}
\end{figure}

Given an image, we form two partitions of the image, one partition involving ``large'' blocks and one involving ``small'' blocks. 
The small blocks are typically one-half the size of the large blocks.  Figure \ref{fig:block_ifs} illustrates this, where the blocks have been
made large enough to be seen clearly.  In a real implementation the blocks would be much smaller.

Given these two block partitions of the image, the \emph{fractal block encoding} algorithm works by scanning through
all the small blocks and, for each such small block, searching amongst the large blocks for the best match. 
The likelihood of finding a good match for all of the small blocks is not very high.  To compensate, we are allowed
to modify the large block by shifting the value of the entire block by a constant, $\beta$, and also scaling each pixelvalue by another constant $\alpha$.
  Figure \ref{fig:block_ifs} indicates the mapping of a large block to a corresponding small block. The algorithm is:

\begin{algorithmic}[1]
    \FOR{ {\bf SB} in small blocks}

        \FOR{ {\bf LB} in large blocks}

            \STATE Downsample {\bf LB} to the same size as {\bf SB}
            \STATE Use leastsquares to find the best parameters $\alpha$ and $\beta$ for this
                    combination of {\bf LB} and {\bf SB}. That is, minimize $\| {\bf SB}  - (\alpha {\bf LB} + \beta)\|_2$.
            \STATE Compute an error for these parameters.  If the error is smaller than for any other {\bf LB}, remember this pair
                   along with the $\alpha$ and $\beta$.
         \ENDFOR

    \ENDFOR
\end{algorithmic}

At the end of this procedure, for each small block we have found an optimally matching large block along
with the $\alpha$ and $\beta$ parameters for the match.  This list of triples (\emph{index of the large block}, $\alpha$, $\beta$)
forms the encoding of the image.  
The scheme essentially uses the $\beta$ parameters to store a coarse version of the image and then extrapolates the fine detail in the image from this coarse image by using the $\alpha$
parameters along with the choice of which parent block matched a given child block.

\begin{figure}[htb]
\begin{center}
    \includegraphics[width = 2 in]{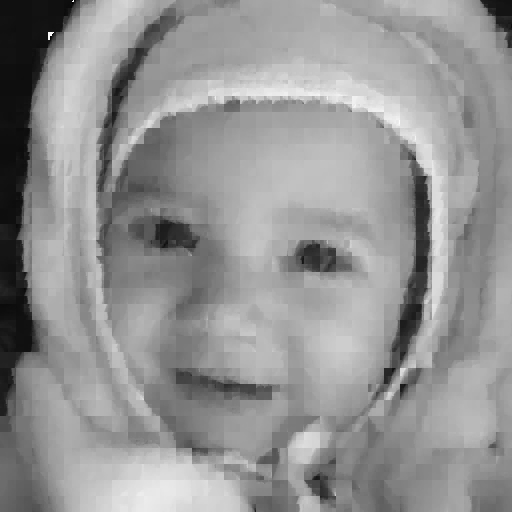} \hskip 1 cm
    \includegraphics[width = 2 in]{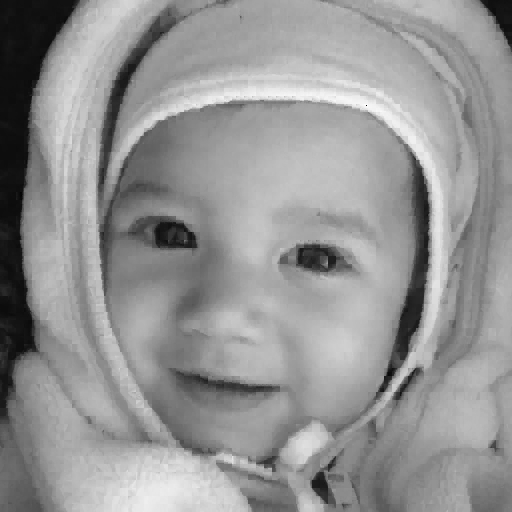}
  \caption{Results of standard Fractal block compression with two choices of the size of the ``small'' blocks," $16 \times 16$ for the first image and $8 \times 8$ for the second.}
  \label{fig:FC}
\end{center}
\end{figure}

In Figure \ref{fig:FC} we see the results of this algorithm on our test image with two choices of the size of the ``small'' blocks.
We used 4 bits to represent $\alpha$ and  9 bits for $\beta$.  In the first image, there are $256$ large blocks so we need $8$ bits for the index.
In the second  image, there are $1024$ large blocks so we need $10$ bits for the index.  Thus the first image is stored in $1024( 4 + 9 + 8) = 2688\mbox{B}$ and the
second is stored in $4096(4+9+10) = 11776\mbox{B}$.

Both the standard Fractal block coding method and our novel V-variable algorithm can be viewed abstractly as some type of block vector quantization algorithm  where the code book is constructed from the image itself \cite{SHH}.  However, our V-variable algorithm has the benefit that this process is repeated independently at all scales, whereas the standard Fractal block coding algorithm only does  this once at the scale of the ``small'' blocks.

\begin{figure}[htp]
 \begin{center}
    \begin{tabular}{cc}
\includegraphics[width=2.0in]{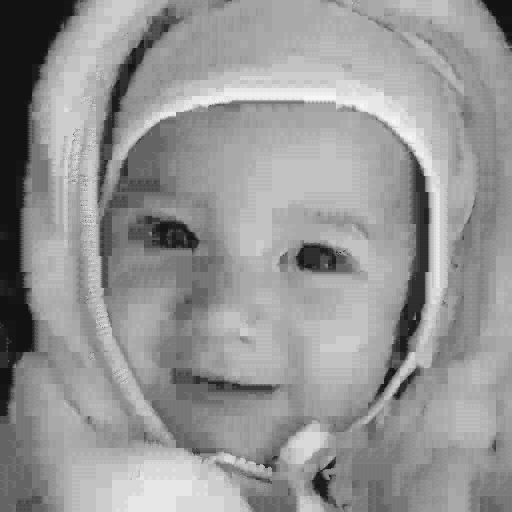} &
\includegraphics[width=2.0in]{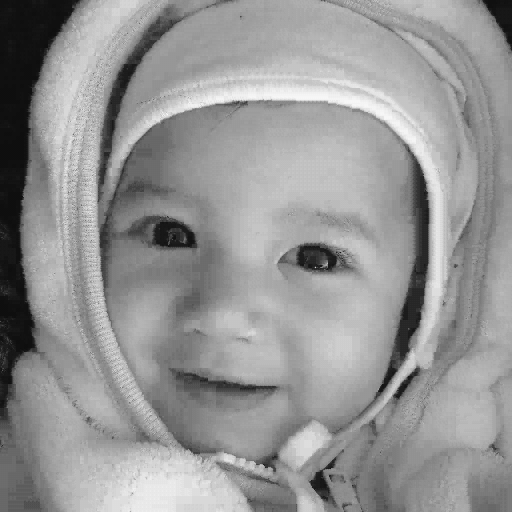} \\
\includegraphics[width=2.0in]{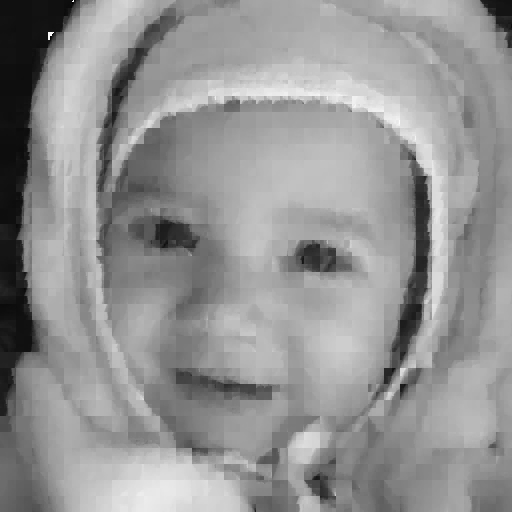} &
\includegraphics[width=2.0in]{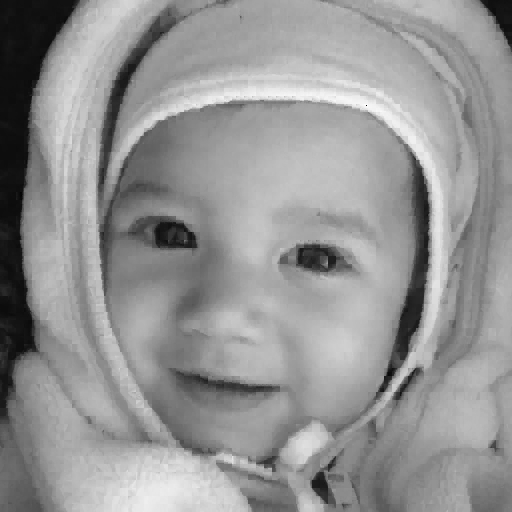} \\
\end{tabular}
\end{center}
\caption{Comparing the approximations 
generated by our V-variable fractal method and  images using different block sizes in
the standard fractal method.  
The top two images are the reconstructions using $V = 256$ and $V = 1024$, respectively.  The bottom two images are obtained from the standard Fractal block method using ``small'' block sizes of $16 \times 16$ and $8 \times 8$, respectively.
The PSNR values for the four images are 27.72, 31.61, 25.84, 29.76, respectively.}
\label{fig:Comparison}
\end{figure}

In Figure \ref{fig:Comparison} we compare the results of our novel V-variable algorithm and the standard Fractal bock coding method.  
The top two images in this figure are the reconstructions using $V = 256$ and $V = 1024$, respectively.  
The bottom two images are obtained from the standard Fractal block method using ``small'' block sizes of $16 \times 16$ and $8 \times 8$, respectively.  
As mentioned previously, the compressed sizes are $5120\mbox{B}$ and $19456\mbox{B}$ for the two top images and $2688\mbox{B}$ and $11776\mbox{B}$ for the bottom two.

\begin{figure}[h]   
\begin{center}
\begin{tabular}{c}
\includegraphics[width=4.0in]{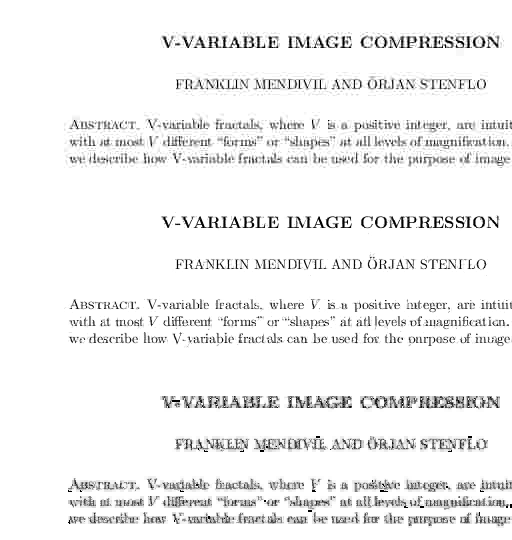}
\end{tabular}
\caption{A comparison between JPEG (upper image), our V-variable method
(middle image), and the standard fractal block method (lower image).
 All images are parts of images  generated as approximations to  a given $512 \times 512$ grayscale textimage.
By using our method we can avoid the  ``halos''
appearant in the JPEG image. The standard fractal method is not competitive
 here. Lossless compression, like e.g.\ PNG, requires larger file sizes.}
 \label{fig:text} 
\end{center}
\end{figure}

In Figure \ref{fig:text} we compare the results of a  16K JPEG encoding, a 1024-variable encoding, and a standard fractal block method encoding using $4 \times 4$ block size
of a given text image.
The JPEG image was generated by saving the given grayscale textimage 
as a  JPEG file with quality level 10 using GIMP (The GNU Image Manipulation Program).
Note that such a storing typically involves  a visually almost lossless reduction of the colour space and a code close to being opimally compressed while the $1024$-variable code described here is far from being optimal.

\section{Generalisations} 
\label{general}

The V-variable image compression algorithm can be generalised in a variety of ways.
We may for example let the number of distinct image pieces vary from level to level, use IFSs other than the IFS in Example \ref{ex:3} as long as its attractor forms a tiling of the unit square,
introduce parameters playing the same role as the $\alpha$ and $\beta$ values in the fractal block method, 
 and use hybrid methods 
combining V-variable and wavelet techniques.
 
The efficiency of our algorithm depends crucially on the clustering method we use. 
In our implementation in Matlab we used  {\tt kmeans} for convenience since it is built-in. 
Matlab also supports ``hierarchical clustering'' which could have been another simple alternative. 
There are indeed many approaches to clustering;  exploring these approaches will provide us with an interesting future problem that may lead to higher compression ratios and improved computational efficiency.
In particular, the methods used in vector quantization for constructing the codebook could be explored.
Another future possibility is a hybrid between our V-variable approach and trellis-coded quantization \cite{SA}.

\bigskip

\newpage

\begin{center}
 {\bf Acknowledgements}
\end{center} 
We are grateful to Jimmy Azar and Cris Luengo for  helpful discussions.  F. Mendivil is partially supported by a grant from the Natural Sciences and Engineering Research Council of Canada (NSERC).

\medskip

\end{document}